\documentclass[letterpaper]{article}
\usepackage[latin9]{inputenc}
\usepackage{verbatim}
\usepackage{amsmath}
\usepackage{amssymb}
\usepackage{graphicx}

\makeatletter

\providecommand{\tabularnewline}{\\}

\usepackage{spconf}
\twoauthors{Fred Richardson, Douglas Reynolds}
 {MIT Lincoln Laboratory\\
  Lexington, MA}
 {Najim Dehak}
  {MIT CSAIL\\
  Cambridge, MA}

\makeatother

\begin{document}

\title{A Unified Deep Neural Network for Speaker and Language Recognition%
\thanks{{\footnotesize{}This work was sponsored by the Department of Defense
under Air Force contract FA8721-05-C-0002. Opinions, interpretations,
conclusions, and recommendations are those of the authors and are
not necessarily endorsed by the United States Government.}%
}}
\maketitle
\begin{abstract}
\global\long\def\x{\mathbf{x}}
\global\long\def\a{\mathbf{a}}
\global\long\def\h{\mathbf{h}}
\global\long\def\W{\mathbf{W}}
\global\long\def\M{\mathbf{M}}
\global\long\def\T{\mathbf{T}}
\global\long\def\WC{\Sigma_{\mathrm{wc}}}
\global\long\def\AC{\Sigma_{\mathrm{ac}}}
\global\long\def\m{\mathbf{m}}
Learned feature representations and sub-phoneme posteriors from Deep
Neural Networks (DNNs) have been used separately to produce significant
performance gains for speaker and language recognition tasks. In this
work we show how these gains are possible using a single DNN for both
speaker and language recognition. The unified DNN approach is shown
to yield substantial performance improvements on the the 2013 Domain
Adaptation Challenge speaker recognition task (55\% reduction in EER
for the out-of-domain condition) and on the NIST 2011 Language Recognition
Evaluation (48\% reduction in EER for the 30s test condition).
\end{abstract}

\section{Introduction}

The impressive gains in performance obtained using deep neural networks
(DNNs) for automatic speech recognition (ASR) \cite{GHintonIspm2012}
have motivated the application of DNNs to other speech technologies
such as speaker recognition (SR) and language recognition (LR) \cite{YSongIlet2013,PMatejkaOdy2014,ILMorenoIcsp2014,TYamadaIsp2013,YLeiIcsp2014,YLeiOdy2014,PKennyOdy2014,OGhahabiOdy2014}.
Two general methods of applying DNN's to the SR and LR tasks have
been shown to be effective. The first or ``direct'' method uses
a DNN trained as a classifier for the intended recognition task. In
the direct method the DNN is trained to discriminate between speakers
for SR \cite{TYamadaIsp2013} or languages for LR \cite{ILMorenoIcsp2014}.
The second or ``indirect'' method uses a DNN trained for a different
purpose to extract data that is then used to train a secondary classifier
for the intended recognition task. Applications of the indirect method
have used a DNN trained for ASR to extract frame-level features \cite{YSongIlet2013,PMatejkaOdy2014,ASarkarISPL2014},
accumulate a multinomial vector \cite{YLeiOdy2014} or accumulate
multi-modal statistics \cite{YLeiIcsp2014,PKennyOdy2014} that were
then used to train an i-vector system \cite{DehakAsp2011,DehakIsp2011}.

The unified DNN approach described in this work uses two of the indirect
methods described above. The first indirect method (``bottleneck'')
uses frame-level features extracted from a DNN with a special bottleneck
layer \cite{YZhangIcsp2014} and the second indirect method (\textquotedblleft DNN-posterior\textquotedblright )
uses posteriors extracted from a DNN to accumulate multi-modal statistics
\cite{YLeiIcsp2014}. The features and the statistics from both indirect
methods are then used to train four different i-vector systems: one
for each task (SR and LR) and each method (bottleneck and DNN-posterior).
A key point in the unified approach is that a single DNN is used for
all four of these i-vector systems. Additionally, we will examine
the feasibility of using a single i-vector extractor for both SR and
LR.

\section{I-vector classifier for SR and LR}

Over the past 5 years, state-of-the-art SR and LR performance has
been achieved using i-vector based systems \cite{DehakAsp2011}. In
addition to using an i-vector classifier as a baseline approach for
our experiments, we will also show how phonetic-knowledge rich DNN
feature representations and posteriors can be incorporated into the
i-vector classifier framework providing significant performance improvements.
In this section we provide a high-level description of the i-vector
approach (for a detailed description see, for example, \cite{DehakAsp2011,DGRomeroIsp2011}). 

\begin{figure}
\centering{}\includegraphics[scale=0.55]{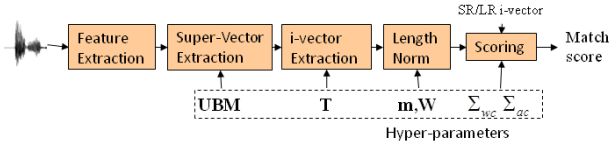}\vspace{-0.15in}
\protect\caption{\label{fig:Simplified-block-diagram}Simplified block diagram of i-vector
extraction and scoring.}
\vspace{-0.15in}
\end{figure}

In Figure \ref{fig:Simplified-block-diagram} we show a simplified
block diagram of i-vector extraction and scoring. An audio segment
is first processed to find the locations of speech in the audio (speech
activity detection) and to extract acoustic features that convey speaker/language
information. Typically 20 dimensional mel-frequency cepstral coefficients
(MFCC) and derivatives are used for SR and 56 dimensional static cepstra
plus shifted-delta cepstra (SDC) are used for LR analyzed at 100 feature
vectors/second. Using a Universal Background Model (UBM), essentially
a speaker/language-independent Gaussian mixture model (GMM), the per-mixture
posterior probability of each feature vector (``GMM-posterior'')
is computed and used, along with the feature vectors in the segment,
to accumulate zeroth, first, and second order sufficient statistics
(SS). These SSs are then transformed into a low dimensional i-vector
representation (typically 400-600 dimensions) using a total variability
matrix, $\T$. The i-vector is whitened by subtracting a global mean,
$\m$, scaled by the inverse square root of a global covariance matrix,
$\W$, and then normalized to unit length \cite{DGRomeroIsp2011}.
Finally, a score between a model and test i-vector is computed. The
simplest scoring function is the cosine distance between the i-vector
representing a speaker/language model (average of i-vectors from the
speaker's/language's training segments) and the i-vector representing
the test segment. The current state-of-the-art scoring function, called
Probabilistic Linear Discriminant Analysis (PLDA) \cite{DGRomeroIsp2011},
requires a within-class matrix $\WC$, characterizing how i-vectors
from a single speaker/language vary, and an across class matrix $\AC$,
characterizing how i-vectors between different speakers/languages
vary. 

Collectively, the UBM, $\T$, $\W$, $\m$, $\WC$, and $\AC$ are
known as the system\textquoteright s hyper-parameters and must be
estimated before a system can enroll and/or score any data. The UBM,
$\T$, $\W$, and $\m$ represent general feature distributions and
total variance of statistics and i-vectors, so unlabeled data from
the desired audio domain (i.e., telephone, microphone, etc.) can be
used to estimate them. The $\WC$ and $\AC$ matrices, however, each
require a large collection of labeled data for training. For SR, $\WC$
and $\AC$ typically require thousands of speakers each of whom contributes
tens of samples to the data set. For LR, the enrollment samples from
each desired languages, which typically hundreds of samples from many
different speakers, can be used to estimate $\WC$ and $\AC$. 

By far the most computationally expensive part of an i-vector system
is extracting the i-vectors themselves. An efficient approach for
performing both SR and LR on the same data is to use the same i-vectors.
This may be possible if both systems use the same feature extraction,
UBM, and $\T$ matrices. There may be some tradeoff in performance
however since the UBM, $\T$ matrix, and signal processing will not
be specialized for SR or LR.

\section{Deep Neural Network Classifier for Speech Applications}

\begin{figure}
\centering{}\includegraphics[scale=0.5]{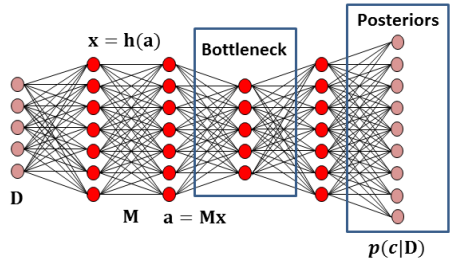}\vspace{-0.15in}
\protect\caption{\label{fig:Example-DNN-architecture}Example DNN architecture}
\vspace{-0.15in}
\end{figure}

\subsection{DNN architecture}

A DNN, like a multi-layer preceptron (MLP), consists of an input layer,
several hidden layers and an output layer. Each layer has a fixed
number of nodes and each sequential pair of layers are fully connected
with a weight matrix. The activations of nodes on a given layer are
computed by transforming the output of the previous layer with the
weight matrix: $\a^{(i)}=\M^{(i)}\x^{(i-1)}$. The output of a given
layer is then computed by applying an ``activation function'' $\x^{(i)}=\h^{(i)}(\a^{(i)})$
(see Figure \ref{fig:Example-DNN-architecture}). Commonly used activation
function include the sigmoid, the hyperbolic tangent, rectified linear
units and even a simple linear transformation. Note that if all the
activation functions in the network are linear then the stacked matrices
reduce to a single matrix multiply. 

The type of activation function used for the output layer depends
on what the DNN is used for. If the DNN is trained as a regression
the output activation function is linear and the objective function
is the mean squared error between the output and some target data.
If the DNN is trained as a classifier then the output activation function
is the soft-max and the objective function is the cross entropy between
the output and the true class labels. For a classifier, each output
node of the DNN classifier correspond to a class and the output is
an estimate of the posterior probability of the class given the input
data.\vspace{-0.2in}

\subsection{DNN Training for ASR}

DNN classifiers can be used as acoustic models in ASR systems to compute
the posterior probability of a sub-phonetic unit (a ``senone'')
given an acoustic observation. Observations, or feature vectors, are
extracted from speech data at a fixed sample rate using a spectral
technique such as filterbank analysis, MFCC, or perceptual linear
prediction (PLP) coefficients. Decoding is preformed using a hidden
Markov model (HMM) and the DNN to find the most likely sequence of
senones given the feature vectors (this requires using Bayes' rule
to convert the DNN posteriors to likelihoods). Training the DNN requires
a significant amount of manually transcribed speech data~\cite{GHintonIspm2012}.
The senones labels are derived from the transcriptions using a phonetic
dictionary and a state-of-the-art GMM/HMM ASR system. Generally speaking,
a refined set of phonotactic units aligned using a high performing
ASR system is required to train a high performing DNN system~\cite{GHintonIspm2012}%
.

DNN training is essentially the same as traditional MLP training.
The most common approach uses stochastic gradient descent (SGD) with
a mini-batch for updating the DNN parameters throughout a training
pass or ``epoch''. The back-propegation algorithm is used to estimate
the gradient of the DNN parameters for each mini-batch. Initializing
the DNN is critical, but it has been shown that a random initialization
is adequate for speech applications where there is a substantial amount
of data~\cite{DengIcsp2013}. A held out validation data set is used
to estimate the error rate after each training epoch. The SGD algorithm
uses a heuristic learning rate parameter that is adjusted in accordance
with a scheduling algorithm which monitors the validation error rate
at each epoch. Training ceases when the error rate can no longer be
reduced.

In the past, training neural networks with more than 2 hidden layers
proved to be problematic. Recent advances in fast and affordable computing
hardware, optimization software and initialization techniques have
made it possible to train much deeper networks. A typical DNN for
ASR will have 5 or more hidden layers each with the same number of
nodes - typically between 500 and 3,000~\cite{GHintonIspm2012}.
The number of output senones varies from a few hundred to tens of
thousands~\cite{DengIcsp2013}.\vspace{-0.1in}

\subsection{DNN bottleneck features}

A DNN can also be used as a means of extracting features for use by
a secondary classifier - including another DNN \cite{VeselyASRU2011}.
This is accomplished by sampling the activation of one of the DNN's
hidden layers and using this as a feature vector. For some classifiers
the dimensionality of the hidden layer is too high and some sort of
feature reduction is necessary like LDA or PCA. In \cite{YZhangIcsp2014},
a dimension reducing linear transformation is optimized as part of
the DNN training by using a special bottleneck hidden layer that has
fewer nodes (see Figure \ref{fig:Example-DNN-architecture}). The
bottleneck layer uses a linear activation so that it behaves very
much like a LDA or PCA transformation on the activation of the previous
layer. The bottleneck DNN used in this work is the same system described
in \cite{YZhangIcsp2014}. In theory any layer can be used as a bottleneck
layer, but in our work we have chosen to use the second to last layer
with the hope that the output posterior prediction will not be too
adversely affected by the loss of information at the bottleneck.\vspace{-0.1in}

\subsection{DNN stats extraction for an i-vector system}

A typical i-vector system uses zeroth, first and second order statistics
generated using a GMM. Statistics are accumulated by first estimating
the posterior of each GMM component density for a frame (the ``occupancy'')
and using these posteriors as weights for accumulating the statistics
for each component of the mixture distribution. The zeroth order statistics
are the total occupancies for an utterance across all GMM components
and the first order statistics are the weighted sum of the means per
a component. The i-vector is then computed using a dimension reducing
transformation that is non-linear with respect to the zeroth order
statistics. 

An alternate approach to extracting statistics has been proposed in~\cite{YLeiIcsp2014}.
Statistics are accumulated in the same way as for the GMM but class
posteriors from the DNN are used in place of GMM component posteriors.
Once the statistics have been accumulated, the i-vector extraction
is performed in the same way as it is from the GMM based statistics.
This approach has been shown to give significant gains for both SR
and LR~\cite{YLeiIcsp2014,YLeiOdy2014,DGRomeroSlt2014}.\vspace{-0.1in}

\section{Experiment setup}

\subsection{Corpora}

Three different corpora are used in our experiments. The DNN itself
is trained using a 100 hours subset of Switchboard 1 \cite{Godfrey1992}.
The 100 hour Switchboard subset is defined in the example system distributed
with Kaldi \cite{Povey_ASRU2011}. The SR systems were trained and
evaluated using the 2013 Domain Adaptation Challenge (DAC13) data
\cite{ShumOdy2014}. The LR systems were evaluated on the NIST 2011
Language Recognition Evaluation (LRE11) data \cite{NIST2011LRE}.
Details on the LR training and development data can be found in \cite{ESingerOdy2011}.\vspace{-0.1in}

\subsection{System configuration}

\subsubsection{Commonalities}

All systems use the same speech activity segmentation generated using
a GMM based speech activity detector (GMM SAD). The i-vector system
uses MAP and PPCA to estimate the $\T$ matrix. Scoring is performed
using PLDA \cite{DGRomeroIsp2011}. With the exception of the input
features or multi-modal statistics, the i-vector systems are identical
and use a 2048 component GMM UBM and a 600 dimensional i-vector subspace.
All LR systems use the discriminative backend described in \cite{ESingerOdy2011}.\vspace{-0.1in}

\subsubsection{Baseline systems}

The front-end feature extraction for the baseline LR system uses 7
static cepstra appended with 49 SDC. Unlike the front-end described
in \cite{ESingerOdy2011}, vocal track length normalization (VTLN)
and feature domain nuisance attribute projection (fNAP) are not used.
The front-end for the baseline SR system uses 20 MFCCs including C0
and their first derivatives for a total of 40 features.\vspace{-0.1in}

\subsubsection{DNN system}

The DNN was trained using 4,199 state cluster (``senone'') target
labels generated using the Kaldi Switchboard 1 ``tri4a'' example
system \cite{Povey_ASRU2011}. The DNN front-end uses 13 Gaussianized
PLP coefficients and their first and second order derivatives (39
features) stacked over a 21 frame window (10 frames to either side
of the center frame) for a total of 819 input features. The GMM SAD
segmentation is applied to the stacked features.

The DNN has 7 hidden layers of 1024 nodes each with the exception
of the 6\textsuperscript{th} bottleneck layer which has 64 nodes.
All hidden layers use a sigmoid activation function with the exception
of 6\textsuperscript{th} layer which is linear\cite{YZhangIcsp2014}.
The DNN training is preformed on an nVidia Tesla K40 GPU using custom
software developed at MIT/CSAIL.\vspace{-0.1in}

\section{Experiment Results}

\vspace{-0.1in}

\subsection{Speaker recognition experiments}

Two sets of experiments were run on the DAC13 corpora: ``in-domain''
and ``out-of-domain''. For both sets of experiments, the UBM and
$\T$ hyper-parameters are trained on Switchboard (SWB) data. The
other hyper-parameters (the $\W$, $\m$ , $\WC$ and $\AC$) are
trained on 2004-2008 speaker recognition evaluation (SRE) data for
the in-domain experiments and SWB data for the out-of-domain experiments
(see \cite{ShumOdy2014} for more details). Tables \ref{tab:In-domain-DAC13-results}
and \ref{tab:out-of-domain-DAC13-results} summarize the results for
the in-domain and out-of-domain experiments with the first row of
each table corresponding to the baseline system. While the DNN-posterior
technique with MFCCs gives a significant gain over the baseline system
for both sets of experiments, as also reported in \cite{YLeiIcsp2014}and
\cite{DGRomeroSlt2014}, an even greater gain is realized using bottleneck
features with a GMM. Unfortunately, using both bottleneck features
and DNN-posteriors degrades performance.\vspace{-0.1in}

\begin{table}
\begin{centering}
\begin{tabular}{|c|c|c|c|}
\hline 
Features & Posteriors & EER(\%) & DCF{*}1000\tabularnewline
\hline 
\hline 
MFCC & GMM & 2.71 & 0.404\tabularnewline
\hline 
MFCC & DNN & 2.27 & 0.336\tabularnewline
\hline 
Bottleneck & GMM & \textbf{2.00} & \textbf{0.269}\tabularnewline
\hline 
Bottleneck & DNN & 2.79 & 0.388\tabularnewline
\hline 
\end{tabular}\vspace{-0.05in}

\par\end{centering}

\protect\caption{\label{tab:In-domain-DAC13-results}In-domain DAC13 results}
\vspace{-0.1in}
\end{table}

\begin{table}
\centering{}%
\begin{tabular}{|c|c|c|c|}
\hline 
Features & Posteriors & EER(\%) & DCF{*}1000\tabularnewline
\hline 
\hline 
MFCC & GMM & 6.18 & 0.642\tabularnewline
\hline 
MFCC & DNN & 3.27 & 0.427\tabularnewline
\hline 
Bottleneck & GMM & \textbf{2.79} & \textbf{0.342}\tabularnewline
\hline 
Bottleneck & DNN & 3.97 & 0.454\tabularnewline
\hline 
\end{tabular}\vspace{-0.05in}
\protect\caption{\label{tab:out-of-domain-DAC13-results}Out-of-domain DAC13 results}
\vspace{-0.1in}
\end{table}

\subsection{Language recognition experiments\vspace{-0.01in}
}

The experiments run on the LRE11 task are summarized in Table \ref{tab:LRE11-results}
with the first row corresponding to the baseline system and the last
row corresponding to a fusion of 5 ``post-evaluation'' systems (see
\cite{ESingerOdy2011} for details). Bottleneck features with GMM
posteriors out performs the other systems configurations including
the 5 system fusion. Interestingly, bottleneck features with DNN-posteriors
show more of an improvement over the baseline system than in the speaker
recognition experiments.\vspace{-0.1in}

\begin{table}
\begin{centering}
\begin{tabular}{|c|c|c|c|c|}
\hline 
Features & Posteriors & 30s & 10s & 3s\tabularnewline
\hline 
\hline 
SDC & GMM & 5.26 & 10.7 & 20.9\tabularnewline
\hline 
SDC & DNN & 4.00 & 8.21 & 19.5\tabularnewline
\hline 
Bottleneck & GMM & \textbf{2.76} & \textbf{6.55} & \textbf{15.9}\tabularnewline
\hline 
Bottleneck & DNN & 3.79 & 7.71 & 18.2\tabularnewline
\hline 
\multicolumn{2}{|c|}{5-way fusion} & 3.27 & 6.67 & 17.1\tabularnewline
\hline 
\end{tabular}\vspace{-0.05in}

\par\end{centering}

\protect\caption{\label{tab:LRE11-results}LRE11 results $C_{avg}$ }
\vspace{-0.1in}
\end{table}

\subsection{Cross-task i-vector Extraction}

Table \ref{tab:Cross-task-DNN-BNF-I-vector} shows the performance
on the DAC13 and LRE11 tasks when extracting i-vectors using parameters
from one of the two systems. As expected, there is a degradation in
performance for the mis-matched task, but the degradation is less
on the DAC13 SR task using the LRE11 LR hyper-parameters. These result
motivate further research in developing a unified i-vector extraction
system for both SR and LR by careful UBM/T training data selection.
\vspace{-0.1in}

\begin{table}
\centering{}%
\begin{tabular}{|c|c|c|}
\hline 
UBM/$\T$ & DAC13 in-domain & LRE11 30s\tabularnewline
\hline 
\hline 
DAC13 & 2.00\% EER / 0.269 DCF & 6.12 $C_{avg}$\tabularnewline
\hline 
LRE11 & 2.68\% EER / 0.368 DCF & 2.76 $C_{avg}$\tabularnewline
\hline 
\end{tabular}\vspace{-0.05in}
\protect\caption{\label{tab:Cross-task-DNN-BNF-I-vector}Cross-task DNN-bottelneck
feature i-vector systems}
\vspace{-0.1in}
\end{table}

\section{Conclusions\vspace{-0.1in}
}

This paper has presented a DNN bottleneck feature extractor that is
effective for both speaker and language recognition and produces significant
performance gains over state-of-the-art MFCC/SDC i-vector approaches
as well as more recent DNN-posterior approaches. For the speaker recognition
DAC13 task, the new DNN bottleneck features decreased in-domain EER
by 26\% and DCF by 33\% and out-of-domain EER by 55\% and DCF by 47\%.
The out-of-domain results are particularly interesting since no in-domain
data was used for DNN training or hyper-parameter adaptation. On LRE11,
the same bottleneck features decreased EERs at 30s, 10s, and 3s test
durations by 48\%, 39\%, and 24\%, respectively, and even out performed
a 5 system fusion of acoustic and phonetic based recognizers. A final
set of experiments demonstrated that it may be possible to use a common
i-vector extractor for a unified speaker and language recognition
system. Although not presented here, it was also observed that recognizers
using the new DNN bottleneck features produced much better calibrated
scores as measured by CLLR metrics.

The DNN bottleneck features, in essence, are the learned feature representation
from which the DNN posteriors are derived. Experimentally, it appears
that using the learned feature representation is better than using
just the output posteriors with SR or LR features, but combining the
DNN bottleneck features and DNN posteriors degrades performance. This
may be because we are able to train a better suited posterior estimator
(UBM) with data more matched to the task data. Since we are working
with new features, future research will examine whether there are
more effective classifiers to apply than i-vectors. Other future research
will explore the sensitivity of the bottleneck features to the DNN's
configuration, and training data quality and quantity. \vspace{-0.1in}

\subsubsection*{Acknowledgments\vspace{-0.1in}
}

The authors would like to thank Patrick Cardinal, Yu Zhang and Ekapol
Chuangsuwanich at MIT CSAIL for sharing their DNN expertise and GPU
optimized DNN training software.\vspace{-0.1in}

\bibliographystyle{IEEEbib}
\bibliography{refs}

\end{document}